\documentclass{article} 
\usepackage[preprint]{colm2026_conference}

\usepackage{microtype}
\usepackage{hyperref}
\usepackage{url}
\usepackage{booktabs}
\usepackage{graphicx}
\usepackage{enumitem}
\usepackage{wrapfig}
\usepackage[most]{tcolorbox}
\usepackage{arydshln}
\usepackage[T1]{fontenc}
\usepackage{listings}
\usepackage{upquote}
\lstdefinestyle{promptStyle}{
  basicstyle=\fontsize{8}{9}\fontfamily{pcr}\selectfont,
  showstringspaces=false,
  breaklines=true,
  breakatwhitespace=false,
  breakindent=0pt,
  keepspaces=false,
  showspaces=false,
  escapeinside={(*@}{@*)},
  frame=none,
  aboveskip=0pt,
  belowskip=0pt,
}

\usepackage{lineno}
\usepackage{pifont}

\definecolor{darkblue}{rgb}{0, 0, 0.5}
\hypersetup{colorlinks=true, citecolor=darkblue, linkcolor=darkblue, urlcolor=darkblue}

\title{How Well Do Agentic Skills Work in the Wild: \\Benchmarking LLM Skill Usage in Realistic Settings}

\newcommand{\aspace}{\hspace{0.9em}}

\author{
Yujian Liu${^{1}}$\thanks{Equal contribution.} \aspace Jiabao Ji${^{1*}}$ \aspace Li An${^1}$ \aspace Tommi Jaakkola${^{2}}$\thanks{Equal advising.} \aspace Yang Zhang${^{3\dagger}}$ \aspace Shiyu Chang${^{1\dagger}}$ \\
$^1$UC Santa Barbara \aspace $^2$MIT CSAIL \aspace $^3$MIT-IBM Watson AI Lab \\
\texttt{\{yujianliu,jiabaoji,li\_an,chang87\}@ucsb.edu}, \\
\texttt{tommi@csail.mit.edu},\quad\texttt{yang.zhang2@ibm.com}}

\begin{document}

\ifcolmsubmission
\linenumbers
\fi

\maketitle

\begin{abstract}
Agent skills, which are reusable, domain-specific knowledge artifacts, have become a popular mechanism for extending LLM-based agents, yet formally benchmarking skill usage performance remains scarce. 
Existing skill benchmarking efforts focus on overly idealized conditions, where LLMs are directly provided with hand-crafted, narrowly-tailored task-specific skills for each task, whereas in many realistic settings, the LLM agent may have to search for and select relevant skills on its own, and even the closest matching skills may not be well-tailored for the task.
In this paper, we conduct the first comprehensive study of skill utility under progressively challenging realistic settings, where agents must retrieve skills from a large collection of 34k real-world skills and may not have access to any hand-curated skills.
Our findings reveal that the benefits of skills are fragile: performance gains degrade consistently as settings become more realistic, with pass rates approaching no-skill baselines in the most challenging scenarios.
To narrow this gap, we study skill refinement strategies, including query-specific and query-agnostic approaches, and we show that query-specific refinement substantially recovers lost performance when the initial skills are of reasonable relevance and quality.
We further demonstrate the generality of retrieval and refinement on \textsc{Terminal-Bench 2.0}, where they improve the pass rate of \texttt{Claude Opus 4.6} from 57.7\% to 65.5\%.
Our results, consistent across multiple models, highlight both the promise and the current limitations of skills for LLM-based agents.
Our code is available at \url{https://github.com/UCSB-NLP-Chang/Skill-Usage}.
\end{abstract}

\section{Introduction}
\label{sec:introduction}

LLM-based agents are rapidly transforming how people build software, analyze data, and automate complex workflows~\citep{anthropic2026opus46,openai_gpt54_thinking_2026,gemini31pro_modelcard_2026}.
A key mechanism for extending agent capabilities beyond their training knowledge is the use of \emph{skills}, reusable knowledge artifacts that encode domain-specific workflows, API usage patterns, coding conventions, and best practices in a structured format~\citep{agentskills2026}.
Skills have seen broad adoption across major agent platforms, including Claude Code, Codex, and a growing ecosystem of open-source repositories~\citep{anthropic_claude_code_overview,openai_codex_2025,openclaw2025github}, enabling users to transform general-purpose agents into specialists for tasks ranging from data engineering to web development.

\begin{figure}[t]
\begin{center}
\begin{minipage}[t]{0.495\linewidth}
\centering
\includegraphics[width=\linewidth]{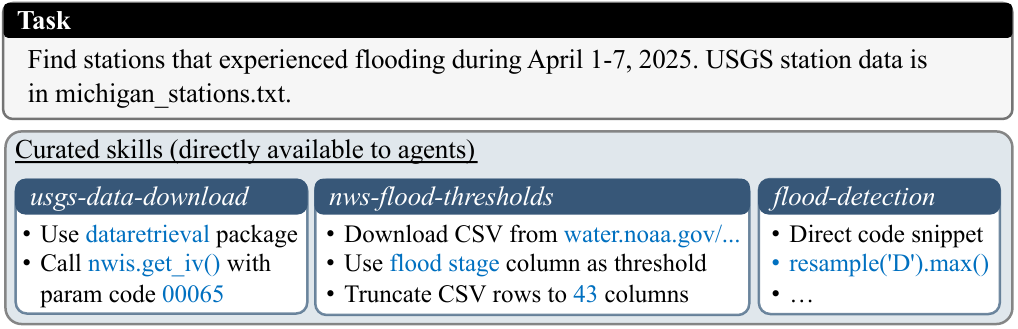}
\end{minipage}
\hfill
\begin{minipage}[t]{0.495\linewidth}
\centering
\includegraphics[width=\linewidth]{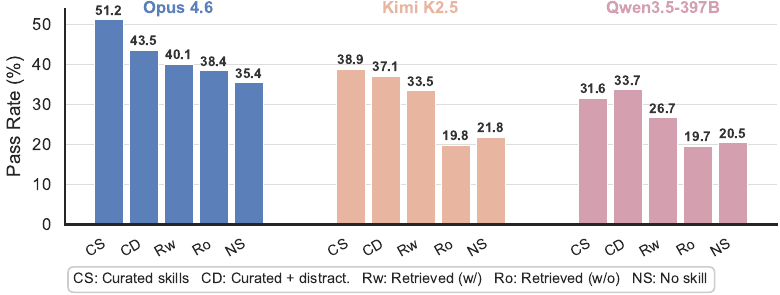}
\end{minipage}
\end{center}
\caption{\textbf{Left:} A \textsc{SkillsBench} example where the task asks agents to identify flooding days at USGS stations. The three curated skills collectively provide the specific API to call, the data source URL for flood thresholds, and code snippets for flood detection (task-specific details are highlighted in blue), effectively forming a step-by-step solution guide. These skills are directly placed in the agent's context without requiring retrieval. \textbf{Right:} Agent pass rates on \textsc{SkillsBench} degrade as evaluation settings become more realistic, from curated skills to settings where agents must retrieve skills from a large collection.}
\label{fig:teaser}
\end{figure}

Despite this widespread adoption, there is surprisingly little rigorous evaluation of whether skills actually help agents solve tasks more effectively.
Recent benchmarks such as \textsc{SkillsBench}~\citep{li2026skillsbenchbenchmarkingagentskills} provide initial evidence that skills can improve agent performance.
However, their evaluation setups are overly idealized in two important ways.
First, the skills provided in \textsc{SkillsBench} are hand-crafted to overfit to each evaluation task, often encoding step-by-step guidance specific to the task rather than general-purpose, reusable knowledge.
For example, as shown in Figure~\ref{fig:teaser} (left), one of \textsc{SkillsBench}'s tasks requires identifying flooding days for USGS stations, and it is paired with three curated skills: one detailing how to download water level data from the USGS API, another specifying the exact URL for NWS flood threshold data, and a third containing code snippets for counting flooding days. These skills combined almost directly spell out the exact solution guide for the task.
Second, the curated skills are directly placed in the agent's context, bypassing the practical challenge of discovering the right skills from a large and noisy collection.
These idealizations raise a fundamental question: \emph{Do skills remain helpful under realistic conditions, where agents must retrieve relevant skills from a large, noisy pool and adapt general-purpose, non-task-specific skills to user queries?}

In this work, we conduct a comprehensive study of skill utility under realistic conditions.
To enable this study, we assemble a collection of 34k real-world skills from open-source repositories, filtered by permissive licenses, quality, and deduplicated.
We explore various search methods for skill retrieval, including keyword, semantic, hybrid, and agentic search, and find that agentic hybrid search, where the agent iteratively formulates queries and evaluates candidate skills, significantly outperforms other approaches.

Building on this infrastructure, we evaluate skill utility on \textsc{SkillsBench} under progressively more realistic settings: from augmenting human-curated skills with distractors, to retrieving from the full skill collection (including the curated skills), to retrieving from a collection where the curated skills have been entirely removed.
Among our findings, a key result is that skill benefits degrade consistently as settings become more realistic (Figure~\ref{fig:teaser}, right), with performance eventually approaching no-skill baselines in the most challenging scenario.
This trend is observed across multiple models, including \texttt{Claude Opus 4.6}~\citep{anthropic2026opus46}, \texttt{Kimi K2.5}~\citep{kimiteam2026kimik25visualagentic}, and \texttt{Qwen3.5-397B-A17B}~\citep{yang2025qwen3technicalreport}.
Our analyses also reveal two bottlenecks limiting skill utility: \ding{182} Agents struggle to determine which skills are worth loading, leaving potentially helpful skills unused; and \ding{183} The content of retrieved skills is often noisy or lacks the precise information needed for the task.

To address these bottlenecks, we study skill refinement strategies to both improve skill selection and distill more useful content from noisy retrieved skills. Specifically, we compare query-specific refinement, where the agent explores and adapts retrieved skills to the target task, and query-agnostic refinement, where skills are improved offline without knowledge of the downstream task.
We find that query-specific refinement is beneficial, substantially recovering lost performance when the initially retrieved skills are of reasonable quality, though gains are more limited when relevant skills are absent from the collection.
Finally, to demonstrate the generality of our approach beyond benchmarks designed for skills, we further evaluate skill retrieval and refinement on \textsc{Terminal-Bench 2.0}~\citep{merrill2026terminalbenchbenchmarkingagentshard}, a general-purpose agent benchmark without human-curated skills, and show that skill retrieval and refinement improve the pass rate from 57.7\% to 65.5\% with \texttt{Claude Opus 4.6}.

To summarize, our contributions are as follows:
\begin{itemize}[leftmargin=1.5em]
    \item We introduce a realistic evaluation framework for agent skills with progressively challenging settings that move beyond the idealized assumptions of prior works, and provide empirical evidence that skill benefits are fragile and degrade under realistic conditions.
    \item We conduct a comprehensive skill retrieval study, comparing keyword, semantic, hybrid, and agentic search strategies, and demonstrate the effectiveness of agentic hybrid search.
    \item We present an in-depth analysis of skill refinement strategies, including query-specific and query-agnostic approaches, revealing when and why refinement helps.
\end{itemize}

\section{Related Work}
\label{sec:related_work}

\paragraph{Reusable knowledge for LLM agents.}
A growing body of work explores how LLM agents can accumulate and reuse knowledge across tasks, taking various forms including programmatic tools and actions~\citep{cai2024largelanguagemodelstool,nguyen2025dynasaur,wang2025inducing}, skill libraries built through exploration in embodied environments~\citep{wang2023voyageropenendedembodiedagent,shi2026evolvingprogrammaticskillnetworks}, structured instruction manuals~\citep{chen2024automanual,liu2025learningonlinevideosinference}, reusable workflows and procedural memory extracted from agent experience~\citep{zhao2024expelllmagentsexperiential,wang2024agentworkflowmemory,mi2026procmemlearningreusableprocedural}, and persistent agent memory that retains useful knowledge across sessions~\citep{hu2026memoryageaiagents}.
Several recent works further study how such knowledge can be automatically evolved and improved over time through self-improvement loops~\citep{zheng2025skillweaverwebagentsselfimprove} or reinforcement learning~\citep{xia2026skillrlevolvingagentsrecursive,wang2026reinforcementlearningselfimprovingagent}.
While these works demonstrate broad interest in reusable knowledge, they each adopt different formats and definitions, and focus primarily on knowledge creation and evolution.
Our work studies a standardized skill format and addresses the complementary question of whether retrieved skills actually help under realistic conditions.

\paragraph{Agentic skills.}
A standardized notion of \emph{agentic skills} has been recently proposed: file-system-based knowledge artifacts consisting of a skill file (\texttt{SKILL.md}) with structured metadata and content, optionally accompanied by helper files~\citep{agentskills2026}.
Following this, a rapidly growing ecosystem of work has emerged around agentic skills, spanning skill taxonomy and lifecycle analysis~\citep{jiang2026sokagenticskills}, large-scale skill infrastructure~\citep{liang2026skillnetcreateevaluateconnect}, automated skill discovery and evolution~\citep{yang2026autoskillexperiencedrivenlifelonglearning,alzubi2026evoskillautomatedskilldiscovery}, skill routing at scale~\citep{zheng2026skillrouterskillroutingllm}, skills as persistent evolving memory~\citep{zhou2026mementoskillsletagentsdesign}, and security risks of third-party skill files~\citep{schmotz2026skillinjectmeasuringagentvulnerability}.
On the benchmarking side, \citet{li2026skillsbenchbenchmarkingagentskills} introduces \textsc{SkillsBench} and \citet{han2026sweskillsbenchagentskillsactually} studies skills in real-world software engineering, but both evaluate under idealized conditions where curated skills are directly provided.
Our work is the first to systematically evaluate skill utility under progressively realistic conditions and to study refinement strategies for narrowing the resulting performance gap.

\paragraph{Agent self-improvement and test-time adaptation.}
Our skill refinement strategies, where the agent explores and adapts retrieved knowledge to the target task, connect to work on agents that improve through experience and test-time adaptation.
Foundational approaches enable agents to learn from task feedback through verbal self-reflection~\citep{shinn2023reflexionlanguageagentsverbal}, policy gradient optimization~\citep{yao2024retroformerretrospectivelargelanguage}, and memory-based online reinforcement learning~\citep{zhou2025mementofinetuningllmagents}.
More recent work accumulates reusable knowledge at inference time, including adaptive strategies and code snippets~\citep{suzgun2025dynamiccheatsheettesttimelearning}, generalizable reasoning patterns~\citep{ouyang2026reasoningbankscalingagentselfevolving}, and continuously evolving memory~\citep{zhang2026liveevoonlineevolutionagentic,zhang2026memskilllearningevolvingmemory}.
\citet{yan2026tidetrajectorybaseddiagnosticevaluation} provides a diagnostic framework for evaluating test-time improvement in agents.
We refer the reader to \citet{fang2025comprehensivesurveyselfevolvingai} and \citet{jiang2026adaptationagenticaisurvey} for broader surveys of self-evolving agents and agent adaptation paradigms.

\section{Skill Usage in Realistic Settings}
\label{sec:realistic}

As illustrated in Figure~\ref{fig:teaser} (left), prior evaluations provide agents with a small set of hand-curated, task-specific skills directly in context.
In real-world usage, however, this idealized setup bypasses three challenges that agents typically face in practice:
\begin{enumerate}[leftmargin=1.5em]
    \item \textbf{Skill selection.} Even when relevant skills are provided to the agent, it must correctly identify which ones are useful and decide to load them, particularly when they appear among many other available skills.
    \item \textbf{Skill retrieval.} Users rarely provide pre-selected skills for every task. Instead, the agent must search through large skill repositories on its own to find potentially relevant ones.
    \item \textbf{Skill adaptation.} When no skills have been specifically authored for the task at hand, the agent must work with retrieved skills that only partially align with the task requirements, extracting useful information from noisy or tangentially relevant content.
\end{enumerate}
We design experiments that progressively introduce these challenges.
To enable this, we first assemble a large-scale skill collection (\S\ref{sec:collection}) and build a retrieval system to search over it (\S\ref{sec:retrieval}), then evaluate agent performance under increasingly realistic settings on \textsc{SkillsBench} (\S\ref{sec:progressive}).

\subsection{Skill Collection}
\label{sec:collection}

To simulate realistic conditions where agents need to search over a large pool and work with skills not narrowly tailored to user queries, we assemble a collection of real-world skills from open-source repositories.
We source skill metadata from two skill aggregation platforms, \texttt{skillhub.club} and \texttt{skills.sh}\footnote{\url{https://www.skillhub.club/} and \url{https://skills.sh/}.}, then download the full skill folder including the \texttt{SKILL.md} file and other helper files from their original GitHub repositories.
We filter by permissive licenses (MIT and Apache 2.0) to ensure redistribution rights, remove ill-formatted skills with empty names or descriptions, and deduplicate by file content.
The resulting collection contains 34,198 skills spanning diverse domains, including web development, data engineering, development operations, scientific computing, etc.

\subsection{Skill Search Engine}
\label{sec:retrieval}

A critical challenge in realistic skill usage is retrieving relevant skills from a large collection. To facilitate the evaluation of the skill retrieval capabilities in LLMs, we build a skill engine tool with a skill index and compare multiple retrieval strategies of increasing sophistication.

\paragraph{Skill index.}
Each skill is indexed with two representations: \ding{172} \emph{metadata}, a concatenation of the skill's name and description, and \ding{173} \emph{full content} in \texttt{SKILL.md}.
We use Qwen3-Embedding-4B~\citep{zhang2025qwen3embeddingadvancingtext} for dense embeddings and BM25 for sparse keyword matching.

\paragraph{Search methods.}
We compare two categories of retrieval approaches:
\begin{itemize}[leftmargin=1.5em]
    \item \textbf{Direct search}: the task description is used as a query to retrieve the top-$k$ skills based on similarity of dense embeddings over the metadata index.
    \item \textbf{Agentic search}: the agent is given access to search tools and iteratively formulates queries, retrieves candidates, and evaluates their relevance before selecting a final set of skills. We evaluate four agentic variants:
    \ding{172} keyword: the agent has access to a BM25-based search tool only;
    \ding{173} semantic: the agent has access to a dense embedding search tool only;
    \ding{174} hybrid w/o content: the agent has access to all three tools (keyword, semantic, and a hybrid tool that combines their scores), with similarity computed over the metadata index only;
    \ding{175} hybrid w/ content: same as \ding{174}, but similarity is a weighted average over both the metadata and full skill content indices.
\end{itemize}
Further details on the index and search implementation are provided in Appendix~\ref{sec:retrieval_details}.

\paragraph{Results.}
We measure retrieval quality using Recall@$k$: the fraction of ground-truth skills that appear in the top-$k$ retrieved results, averaged across all tasks. We consider the manually-curated skills in \textsc{SkillsBench} as the ground-truth for each task. For agentic search, we use \texttt{Claude Opus 4.6} with \texttt{Claude Code} as the agent.

\begin{table}[t]
\centering
\begin{tabular}{lccc}
\toprule
Method & Recall@3 & Recall@5 & Recall@10 \\
\midrule
Direct (semantic) & 38.1 & 47.0 & 52.3 \\
\midrule
Agentic (keyword) & 24.1 & 26.6 & 27.5 \\
Agentic (semantic) & 56.8 & 63.1 & 66.5 \\
Agentic (hybrid) w/o content & \textbf{57.7} & 63.5 & 66.7 \\
Agentic (hybrid) w/ content & 57.3 & \textbf{65.5} & \textbf{68.3} \\
\bottomrule
\end{tabular}
\caption{Skill retrieval performance of \texttt{Claude Opus 4.6} with \texttt{Claude Code} on \textsc{SkillsBench} (Recall@$k$, \%). The retrieval pool contains the curated skills among 34k total skills.}
\label{tab:retrieval}
\end{table}

Table~\ref{tab:retrieval} reports the results. As shown, agentic search substantially outperforms direct search. With the same semantic search tool, agentic search outperforms direct search in Recall@3 by 18.7 points, as the agent can iteratively formulate queries, inspect returned candidates, and refine its search strategy beyond a single fixed query.
Among the agentic variants, using semantic search tool greatly outperforms keyword search tool, indicating that semantic similarity is essential for skill retrieval.
Adding the full skill content index provides a modest but consistent gain at higher $k$ values (Recall@5: 63.5\% $\rightarrow$ 65.5\%; Recall@10: 66.7\% $\rightarrow$ 68.3\%), as the full skill content captures information not covered by metadata alone, enabling broader search over the skill collection.
Based on these results, we use agentic hybrid search with full skill content as the default retrieval method in subsequent experiments.

\subsection{Progressive Evaluation Settings}
\label{sec:progressive}

We now evaluate skill utility on \textsc{SkillsBench} under progressively realistic settings that systematically vary three factors: whether the agent must select which skills to load on its own (forced vs.\ autonomous), how skills are discovered (user-provided vs.\ agent-retrieved), and what skills are available (human-curated vs.\ general-purpose).

\paragraph{Settings.}
We define the following evaluation conditions, ordered from most idealized to most realistic. Each setting introduces one of the three challenges identified above.
\begin{itemize}[leftmargin=1.5em]
    \item \textbf{Curated + forced load}: the original curated skills are placed in the agent's environment, and the agent is explicitly instructed to load all of them. This represents an upper bound on curated skill utility, bypassing all three challenges.
    \item \textbf{Curated}: the original \textsc{SkillsBench} setup, where curated skills are placed in the agent's environment, but whether and when to load them are deferred to the agent itself. This introduces the challenge of \emph{skill selection}: the agent must recognize which available skills are worth loading.
    \item \textbf{Curated + distractors}: all curated skills remain available to the agent, but we add distracting skills retrieved via agentic search from the 34k collection, keeping the total number of skills at 5 for consistency with the retrieval settings. This intensifies the \emph{selection} challenge, as the agent must identify curated skills among noise.
    \item \textbf{Retrieved (w/ curated)}: the agent retrieves top-5 skills from the 34k collection augmented with the curated skills. This introduces the challenge of \emph{skill retrieval}: relevant skills are no longer directly provided, and the agent must search for skills on its own.
    \item \textbf{Retrieved (w/o curated)}: the agent retrieves top-5 skills from the 34k collection without curated skills. This further introduces the challenge of \emph{skill adaptation}: no skills have been specifically authored for the tasks, and the agent must extract useful information from general-purpose skills that only partially align with the task requirements.
    \item \textbf{No skills}: the agent receives no skills, serving as the baseline.
\end{itemize}

\paragraph{Models and evaluation.}
We evaluate with \texttt{Claude Opus 4.6}, \texttt{Kimi K2.5}, and \texttt{Qwen3.5-397B-A17B}, representing frontier proprietary and strong open-weight models. Each model is paired with its native agent harness: Claude Code for Claude, Terminus-2~\citep{harbor2026} for Kimi,\footnote{Terminus-2 was used by Kimi's evaluation on Terminal-Bench 2.0.} and Qwen-Code for Qwen. Each model independently runs the entire pipeline, including skill retrieval, task completion, and later refinement (\S\ref{sec:refinement}), so that results reflect the end-to-end capability of each model and harness pair.
We evaluate on 84 tasks from \textsc{SkillsBench} (excluding tasks with known issues), running each condition 3 times per task. Further details are provided in Appendix~\ref{sec:experiment_details}.

\begin{figure}[t]
\begin{center}
\includegraphics[width=\linewidth]{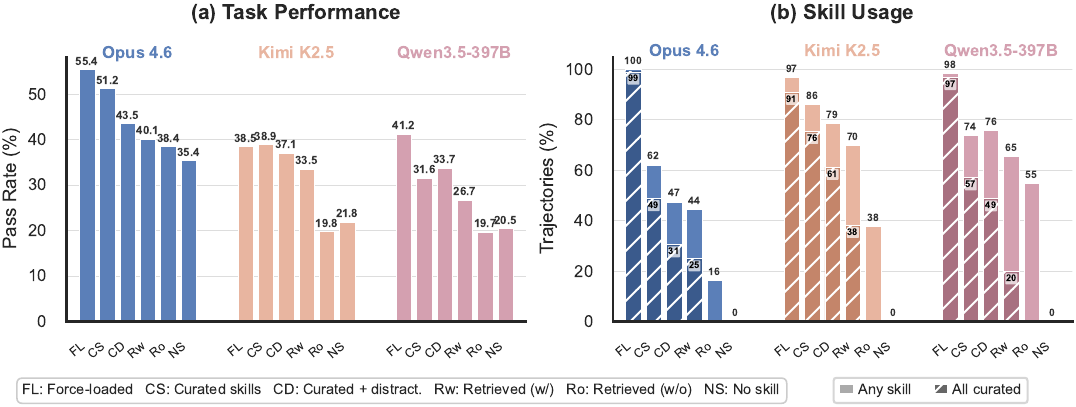}
\end{center}
\caption{\textbf{(a)} Pass rates on \textsc{SkillsBench} under progressively realistic settings, including a force-loaded upper bound. Performance degrades consistently as settings become more realistic. \textbf{(b)} Skill usage across settings. Solid bars show the fraction of trajectories that load any skill; hatched bars show the fraction that load all curated skills. Agents often fail to load curated skills even when they are directly available, and the gap widens as distractors are added and retrieval is required.}
\label{fig:progressive}
\end{figure}

\paragraph{Results.}
Figure~\ref{fig:progressive} presents the main results. Panel (a) shows average pass rates, while panel (b) shows skill usage: the fraction of trajectories that load any skill (solid bars) and the fraction that load all curated skills (hatched bars). We highlight three key observations.

\paragraph{Skill Selection: Agents fail to select the right skills, even when they are directly available.}
The first three settings all provide curated skills to the agent, yet performance drops substantially across them. Force-loading curated skills yields 55.4\% for Claude, but simply letting the agent decide which to load reduces this to 51.2\%, even though the same skills are available. Adding distractors causes a further drop to 43.5\%. The skill usage panel reveals the reason: only 49\% of Claude trajectories load all curated skills in the curated setting, falling to 31\% with distractors. Qwen shows a similar performance pattern (41.2\% $\rightarrow$ 31.6\% $\rightarrow$ 33.7\%). Interestingly, Kimi exhibits much higher skill loading rates even without forced loading (86\% in the curated setting vs.\ 62\% for Claude), indicating that the agent harness significantly influences skill loading behavior. However, this higher loading rate does not translate into better task performance (38.9\% curated vs.\ 38.5\% force-loaded), indicating that skill utility involves not just loading skills but also effectively utilizing their content.

\paragraph{Skill Retrieval: Requiring agents to retrieve skills further degrades performance.}
When relevant skills are no longer directly provided and the agent must retrieve them, performance drops further: Claude falls to 40.1\% and Kimi to 33.5\% when curated skills remain in the retrieval pool. This compounds the selection challenge with imperfect retrieval (our best retrieval achieves 65.5\% Recall@5 in Table~\ref{tab:retrieval}), meaning curated skills are not always among the candidates the agent sees. The skill usage panel also reflects this: Claude's loading rate drops to 44\% under retrieval, compared to 62\% in the curated setting.

\paragraph{Skill Adaptation: Without curated skills, agents struggle to adapt general-purpose skills and approach the no-skill baseline.}
When curated skills are removed from the retrieval pool entirely, the agent can only find general-purpose skills not tailored to the tasks. Claude drops to 38.4\%, only 3.0 points above the no-skill baseline, and skill usage falls to just 16\% of trajectories. The results are more severe for other models: both Kimi (19.8\% vs.\ 21.8\% baseline) and Qwen (19.7\% vs.\ 20.5\% baseline) drop below their no-skill baselines, indicating that irrelevant retrieved skills can actively mislead the agent, \emph{e.g.,} by spending effort loading and following unhelpful instructions that would have been better ignored entirely. This contrast suggests that \textbf{stronger models can better ignore irrelevant skills, while weaker models are more likely to be hurt by low-quality retrieved skills}.

\paragraph{Summary.}
The observed gap between the force-loaded upper bound and the most realistic setting motivates two directions for skill refinement (\S\ref{sec:refinement}).
First, the sharp drop in skill usage even when curated skills are available suggests that agents struggle to recognize relevant skills from their names and descriptions alone, and refining skill metadata may help agents better select which skills to load.
Second, the difficulty of adapting general-purpose skills motivates refining skill content itself to improve clarity and relevance, making retrieved skills more useful in the absence of curated skills. These observations have motivated us to remove these bottlenecks with skill refinement, which is introduced in the next section.

\section{Narrowing the Gap with Skill Refinement}
\label{sec:refinement}

We now investigate whether \emph{skill refinement}, the process of transforming retrieved skills into more useful forms, can recover the lost performance.
We describe two refinement strategies (\S\ref{sec:refine_methods}) and evaluate them on both \textsc{SkillsBench} and \textsc{Terminal-Bench 2.0} (\S\ref{sec:refine_results}).

\subsection{Refinement Strategies}
\label{sec:refine_methods}

We study two strategies for improving skill quality before the agent attempts the task. Full details including prompts are provided in Appendix~\ref{sec:refinement_details}.

\paragraph{Query-agnostic refinement.}
The progressive evaluation in \S\ref{sec:progressive} shows that high-quality curated skills substantially improve agent performance.
A natural aspiration is therefore to improve the entire 34k skill collection offline to approximate curated-level quality.
However, refining all 34k skills is cost-prohibitive, so we instead apply query-agnostic refinement only to the retrieved skills for each task, treating this as an approximation of what a fully improved collection would provide.
To preserve this offline nature, each retrieved skill is refined independently, without knowledge of the target task or other retrieved skills.

We leverage Anthropic's \texttt{skill-creator},\footnote{\url{https://github.com/anthropics/skills/tree/main/skills/skill-creator}.} a meta-skill that encodes best practices for writing effective skills, to drive the improvement process.
For each skill, the model generates synthetic test queries that the skill might be used for, then runs an agent with and without the skill on these queries.
The model compares the two agents' outputs, self-evaluates whether the skill helped or hurt, and uses this feedback to iteratively improve the skill.
Because this computation happens entirely offline, query-agnostic refinement is cheap at inference time and can be applied as a preprocessing step.
However, it has two limitations: it cannot adapt skills to the specific needs of a given task, and because each skill is refined in isolation, it cannot compose information across multiple retrieved skills.

\paragraph{Query-specific refinement.}
To address these limitations, query-specific refinement allows the agent to directly explore the target task before refining.
The agent reads the task instruction, examines all retrieved skills, attempts an initial solution, and self-evaluates correctness (the agent \emph{does not} have access to the ground-truth verifier).
Based on this exploration, the agent reflects on which skills were useful and which were misleading, then composes a refined set of skills tailored to the specific task.
The agent also has access to the \texttt{skill-creator} meta-skill as guidance for writing effective skill metadata and content.
Crucially, unlike query-agnostic refinement, the agent can merge and synthesize across multiple retrieved skills, extracting the relevant portions from each and combining them into a single coherent skill while discarding irrelevant content.
This strategy has high potential but is also more expensive, as it requires a full exploration pass per task at inference time.

\subsection{Results}
\label{sec:refine_results}

We evaluate both refinement strategies on \textsc{SkillsBench} under the retrieved (w/ curated) and retrieved (w/o curated) settings from \S\ref{sec:progressive}.
To assess generalizability, we additionally evaluate on \textsc{Terminal-Bench 2.0}, a widely-used agent benchmark containing 89 tasks spanning system administration, file manipulation, programming challenges, etc.
Unlike \textsc{SkillsBench}, \textsc{Terminal-Bench 2.0} was not designed with skills in mind and has no curated skills, so the agent retrieves from our full skill collection.
Table~\ref{tab:refinement} presents the results.

\begin{table}[t]
\centering
\begin{tabular}{l cc cc cc}
\toprule
& \multicolumn{2}{c}{\texttt{Claude Opus 4.6}} & \multicolumn{2}{c}{\texttt{Kimi K2.5}*} & \multicolumn{2}{c}{\texttt{Qwen3.5-397B-A17B}} \\
\cmidrule(lr){2-3} \cmidrule(lr){4-5} \cmidrule(lr){6-7}
 & Pass & Load & Pass & Load & Pass & Load \\
\midrule
\multicolumn{7}{c}{\textbf{\textsc{SkillsBench}}} \\
\midrule
Curated skills & 51.2 & 62.2 & 38.9 & 86.1 & 31.6 & 73.8 \\
No skills & 35.4 & 0.0 & 21.8 & 0.0 & 20.5 & 0.0 \\
\midrule
Retrieved (w/ curated) & 40.1 & 44.4 & \textbf{33.5} & 69.7 & 26.7 & 65.5 \\
+ Query-specific & \textbf{48.2} & 72.2 & 26.7 & 95.2 & \textbf{30.8} & 75.0 \\
+ Query-agnostic & 42.0 & 32.9 & --- & --- & 26.2 & 68.3 \\
\midrule
Retrieved (w/o curated) & \textbf{38.4} & 16.3 & 19.8 & 37.7 & 19.7 & 54.8 \\
+ Query-specific & 37.9 & 61.1 & \textbf{23.1} & 90.9 & 21.5 & 69.4 \\
+ Query-agnostic & 37.4 & 12.3 & --- & --- & \textbf{24.6} & 53.2 \\
\midrule
\multicolumn{7}{c}{\textbf{\textsc{Terminal-Bench 2.0}}} \\
\midrule
No skills & 57.7 & 0.0 & 46.6 & 0.0 & 44.7 & 0.0 \\
\midrule
Retrieved & 61.4 & 40.8 & 50.6 & 79.0 & 44.2 & 31.1 \\
+ Query-specific & \textbf{65.5} & 74.9 & \textbf{56.2} & 93.6 & \textbf{49.1} & 42.3 \\
+ Query-agnostic & 63.3 & 33.7 & --- & --- & 44.9 & 38.4 \\
\bottomrule
\end{tabular}
\caption{Effect of skill refinement on pass rates (Pass, \%) and skill loading rates (Load, \% of trajectories that load any skill) across \textsc{SkillsBench} and \textsc{Terminal-Bench 2.0}. Query-specific refinement substantially improves both performance and skill adoption when initially retrieved skills are of high relevance. *Kimi's query-agnostic results are omitted because Terminus-2 does not support subagent operations that are needed.}
\label{tab:refinement}
\end{table}

\begin{figure*}[t]
\centering
\includegraphics[width=\linewidth]{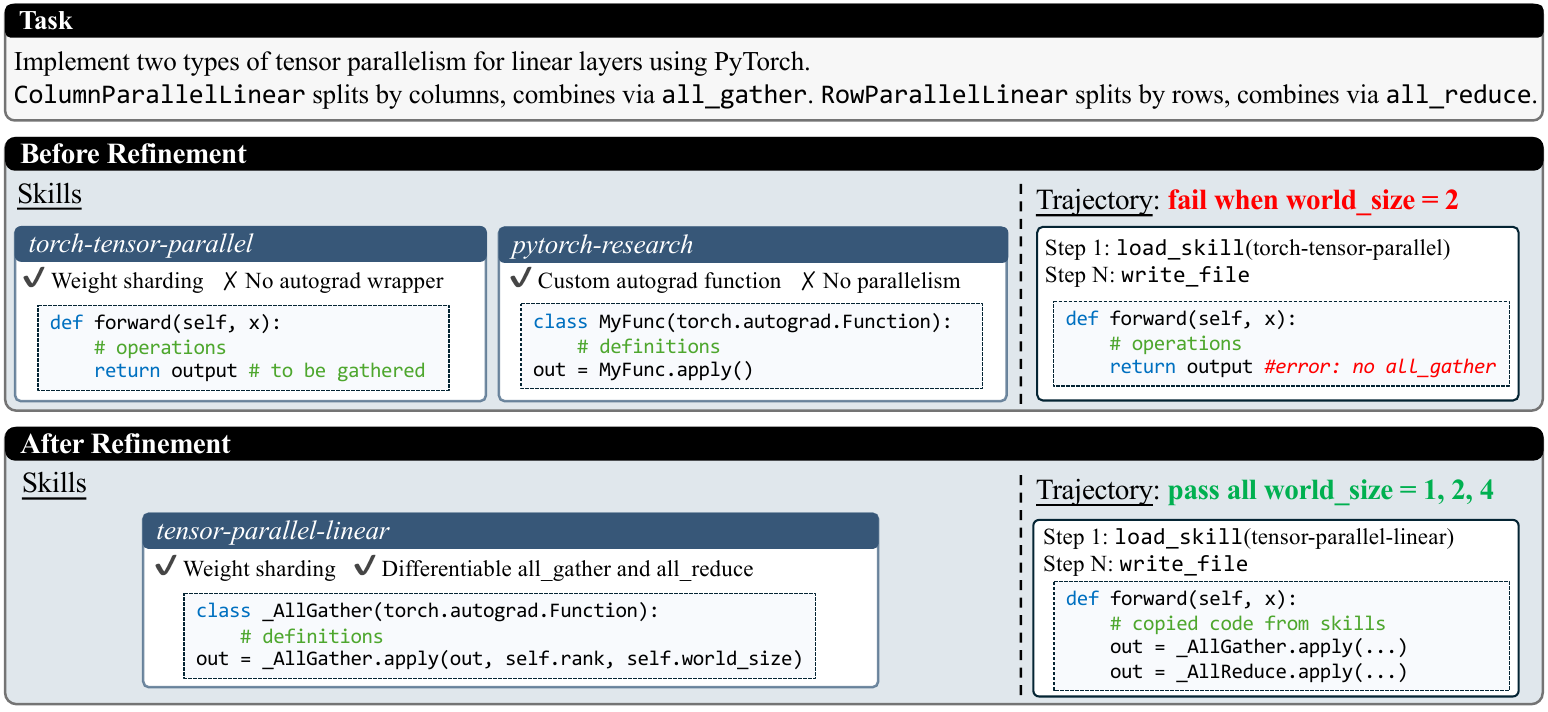}
\caption{Example of query-specific refinement on a \textsc{Terminal-Bench 2.0} tensor parallelism task. \textbf{Top:} Without refinement, the agent retrieves two partially relevant skills but only loads \texttt{torch-tensor-parallel}, ignoring \texttt{pytorch-research}. The loaded skill covers weight sharding but lacks differentiable collective wrappers, leading to wrong implementation for world\_size $>$ 1. \textbf{Bottom:} After refinement, the agent synthesizes a new skill that merges tensor parallelism knowledge from the first skill with custom \texttt{autograd.Function} patterns from the second, producing an implementation that passes all tests.}
\label{fig:refinement_example}
\end{figure*}

\paragraph{Query-specific refinement is broadly effective.}
Query-specific refinement improves performance in 7 out of 9 cases in Table~\ref{tab:refinement}.
On \textsc{SkillsBench} with curated skills in the retrieval pool, it improves Claude from 40.1\% to 48.2\%, recovering most of the gap to the curated setting. For Qwen, the gain is similar: 26.7\% to 30.8\%.
On \textsc{Terminal-Bench 2.0}, where no curated skills exist, query-specific refinement consistently improves all three models: +4.1 for Claude, +5.6 for Kimi, and +4.9 for Qwen, confirming that the benefits extend to a general-purpose benchmark not designed for skills.
The one notable exception is Kimi on \textsc{SkillsBench} w/ curated, where the pass rate drops from 33.5\% to 26.7\%, suggesting that the exploration and self-evaluation process can be counterproductive when the model misjudges which skills are useful.
Notably, skill loading rates also increase substantially with query-specific refinement (\emph{e.g.,} 44\% to 72\% for Claude on \textsc{SkillsBench} w/ curated), indicating that refinement produces skills that agents are more likely to use.
Figure~\ref{fig:refinement_example} illustrates how query-specific refinement composes useful information scattered across multiple retrieved skills: the agent extracts tensor parallelism concepts from one skill and custom autograd patterns from another, synthesizing them into a single skill with differentiable collective operations that neither original skill provides on its own.

\paragraph{Query-agnostic refinement yields smaller gains.}
Query-agnostic refinement provides moderate improvements in some settings (\emph{e.g.,} Claude rises from 40.1\% to 42.0\% on \textsc{SkillsBench} w/ curated and from 61.4\% to 63.3\% on \textsc{Terminal-Bench 2.0}), but the gains are inconsistent and sometimes negligible. Without access to the target task, the improvement process can clean up formatting and improve clarity, but cannot identify which parts of a skill are most relevant or synthesize information across multiple skills. Because query-agnostic refinement moves computation offline, it is cheap at inference time, but the limited and variable gains suggest that task awareness is important for effective refinement.

\paragraph{Refinement effectiveness depends on initial skill quality.}
An interesting pattern in Table~\ref{tab:refinement} is that query-specific refinement yields large gains in some settings but not others.
Under \emph{retrieved (w/o curated)} on \textsc{SkillsBench}, query-specific refinement yields modest or even no gains for three models.
To explain this asymmetry, we assess the relevance and coverage of the initially retrieved skills using an LLM judge (\texttt{GPT-5.4}) that scores each task's retrieved skill set on a 1-5 scale (a higher score means retrieved skills are more relevant and collectively cover different aspects of the target task).

\begin{wraptable}{r}{0.52\textwidth}
\vspace{-1em}
\centering
\begin{tabular}{lccc}
\toprule
Setting & \texttt{Claude} & \texttt{Kimi} & \texttt{Qwen} \\
\midrule
SB: w/ curated & 4.01 & 3.83 & 3.85 \\
SB: w/o curated & 3.49 & 3.31 & 3.39 \\
TB: Retrieved & 4.02 & 3.96 & 4.08 \\
\bottomrule
\end{tabular}
\caption{Average coverage scores of initially retrieved skills, judged by an LLM. Higher scores indicate greater task relevance and coverage.}
\label{tab:coverage}
\vspace{-1em}
\end{wraptable}

Table~\ref{tab:coverage} reveals a clear pattern: the settings where query-specific refinement succeeds (\textsc{SkillsBench} w/ curated, \textsc{Terminal-Bench}) have high initial coverage scores ($\geq$3.83), while the setting where it fails (\textsc{SkillsBench} w/o curated) has notably lower scores ($\leq$3.49).
This confirms that refinement acts more like a \emph{multiplier} on existing skill quality rather than a \emph{generator} of new knowledge.
When the retrieved skills contain relevant information, even if imperfectly matched, query-specific refinement can extract and amplify that signal through exploration and composition.
When relevant skills are absent entirely, it struggles to synthesize useful information.

\section{Conclusion}
\label{sec:conclusion}

We presented a comprehensive study of agent skill utility under realistic conditions, showing that skill benefits degrade substantially as agents must retrieve from large collections and work with general-purpose skills not tailored to the task.
Our further study shows that query-specific refinement can recover much of this lost performance when retrieved skills are of reasonable relevance, but cannot compensate when relevant skills are absent entirely, suggesting that refinement amplifies existing skill quality rather than generating new knowledge.
These findings highlight the need for better skill retrieval, more effective offline refinement methods, and skill ecosystems that account for varying model capabilities.

\section*{Ethics Statement}
This work studies the effectiveness of agent skills under realistic conditions using publicly available benchmarks and open-source skills filtered by permissive licenses (MIT and Apache 2.0).
Our skill collection is sourced from public GitHub repositories and does not contain private or sensitive data.
All model evaluations are conducted on established coding benchmarks in isolated Docker containers, posing no risk to external systems.

\section*{LLM Usage Disclosure}
In accordance with COLM's policy on LLM use, we disclose the following LLM usage.
In research, LLMs were used to assist with modifying existing open-source repositories for the evaluation infrastructure, debugging code, and analyzing agent trajectories.
In writing, LLMs assisted with revising and smoothing text drafted by the authors, proofreading, writing plotting scripts, and formatting tables and other LaTeX elements.
All research ideas, experimental design, and analysis are the work of the authors.

\section*{Acknowledgments}
UCSB acknowledges the support from National Science Foundation(NSF) Grant IIS-2338252, NSF Grant IIS-2302730, and the Open Philanthropy Research Award. Tommi Jaakkola acknowledges the support from NSF Expeditions grant (award 1918839) Understanding the World Through Code.

\newpage

\bibliography{colm2026_conference}

@misc{liu2025learningonlinevideosinference,
      title={Learning from Online Videos at Inference Time for Computer-Use Agents}, 
      author={Yujian Liu and Ze Wang and Hao Chen and Ximeng Sun and Xiaodong Yu and Jialian Wu and Jiang Liu and Emad Barsoum and Zicheng Liu and Shiyu Chang},
      year={2025},
      eprint={2511.04137},
      archivePrefix={arXiv},
      primaryClass={cs.CV},
}

@misc{zheng2024sglangefficientexecutionstructured,
      title={SGLang: Efficient Execution of Structured Language Model Programs}, 
      author={Lianmin Zheng and Liangsheng Yin and Zhiqiang Xie and Chuyue Sun and Jeff Huang and Cody Hao Yu and Shiyi Cao and Christos Kozyrakis and Ion Stoica and Joseph E. Gonzalez and Clark Barrett and Ying Sheng},
      year={2024},
      eprint={2312.07104},
      archivePrefix={arXiv},
      primaryClass={cs.AI},
}

@misc{jiang2026adaptationagenticaisurvey,
      title={Adaptation of Agentic AI: A Survey of Post-Training, Memory, and Skills}, 
      author={Pengcheng Jiang and Jiacheng Lin and Zhiyi Shi and Zifeng Wang and Luxi He and Yichen Wu and Ming Zhong and Peiyang Song and Qizheng Zhang and Heng Wang and Xueqiang Xu and Hanwen Xu and Pengrui Han and Dylan Zhang and Jiashuo Sun and Chaoqi Yang and Kun Qian and Tian Wang and Changran Hu and Manling Li and Quanzheng Li and Hao Peng and Sheng Wang and Jingbo Shang and Chao Zhang and Jiaxuan You and Liyuan Liu and Pan Lu and Yu Zhang and Heng Ji and Yejin Choi and Dawn Song and Jimeng Sun and Jiawei Han},
      year={2026},
      eprint={2512.16301},
      archivePrefix={arXiv},
      primaryClass={cs.AI},
}

@inproceedings{
chen2024automanual,
title={AutoManual: Generating Instruction Manuals by {LLM} Agents via Interactive Environmental Learning},
author={Minghao Chen and Yihang Li and Yanting Yang and Shiyu Yu and Binbin Lin and Xiaofei He},
booktitle={The Thirty-eighth Annual Conference on Neural Information Processing Systems},
year={2024},
}

@misc{hu2026memoryageaiagents,
      title={Memory in the Age of AI Agents}, 
      author={Yuyang Hu and Shichun Liu and Yanwei Yue and Guibin Zhang and Boyang Liu and Fangyi Zhu and Jiahang Lin and Honglin Guo and Shihan Dou and Zhiheng Xi and Senjie Jin and Jiejun Tan and Yanbin Yin and Jiongnan Liu and Zeyu Zhang and Zhongxiang Sun and Yutao Zhu and Hao Sun and Boci Peng and Zhenrong Cheng and Xuanbo Fan and Jiaxin Guo and Xinlei Yu and Zhenhong Zhou and Zewen Hu and Jiahao Huo and Junhao Wang and Yuwei Niu and Yu Wang and Zhenfei Yin and Xiaobin Hu and Yue Liao and Qiankun Li and Kun Wang and Wangchunshu Zhou and Yixin Liu and Dawei Cheng and Qi Zhang and Tao Gui and Shirui Pan and Yan Zhang and Philip Torr and Zhicheng Dou and Ji-Rong Wen and Xuanjing Huang and Yu-Gang Jiang and Shuicheng Yan},
      year={2026},
      eprint={2512.13564},
      archivePrefix={arXiv},
      primaryClass={cs.CL},
}

@misc{cai2024largelanguagemodelstool,
      title={Large Language Models as Tool Makers}, 
      author={Tianle Cai and Xuezhi Wang and Tengyu Ma and Xinyun Chen and Denny Zhou},
      year={2024},
      eprint={2305.17126},
      archivePrefix={arXiv},
      primaryClass={cs.LG},
}

@misc{zhang2025qwen3embeddingadvancingtext,
      title={Qwen3 Embedding: Advancing Text Embedding and Reranking Through Foundation Models}, 
      author={Yanzhao Zhang and Mingxin Li and Dingkun Long and Xin Zhang and Huan Lin and Baosong Yang and Pengjun Xie and An Yang and Dayiheng Liu and Junyang Lin and Fei Huang and Jingren Zhou},
      year={2025},
      eprint={2506.05176},
      archivePrefix={arXiv},
      primaryClass={cs.CL},
}

@misc{merrill2026terminalbenchbenchmarkingagentshard,
      title={Terminal-Bench: Benchmarking Agents on Hard, Realistic Tasks in Command Line Interfaces}, 
      author={Mike A. Merrill and Alexander G. Shaw and Nicholas Carlini and Boxuan Li and Harsh Raj and Ivan Bercovich and Lin Shi and Jeong Yeon Shin and Thomas Walshe and E. Kelly Buchanan and Junhong Shen and Guanghao Ye and Haowei Lin and Jason Poulos and Maoyu Wang and Marianna Nezhurina and Jenia Jitsev and Di Lu and Orfeas Menis Mastromichalakis and Zhiwei Xu and Zizhao Chen and Yue Liu and Robert Zhang and Leon Liangyu Chen and Anurag Kashyap and Jan-Lucas Uslu and Jeffrey Li and Jianbo Wu and Minghao Yan and Song Bian and Vedang Sharma and Ke Sun and Steven Dillmann and Akshay Anand and Andrew Lanpouthakoun and Bardia Koopah and Changran Hu and Etash Guha and Gabriel H. S. Dreiman and Jiacheng Zhu and Karl Krauth and Li Zhong and Niklas Muennighoff and Robert Amanfu and Shangyin Tan and Shreyas Pimpalgaonkar and Tushar Aggarwal and Xiangning Lin and Xin Lan and Xuandong Zhao and Yiqing Liang and Yuanli Wang and Zilong Wang and Changzhi Zhou and David Heineman and Hange Liu and Harsh Trivedi and John Yang and Junhong Lin and Manish Shetty and Michael Yang and Nabil Omi and Negin Raoof and Shanda Li and Terry Yue Zhuo and Wuwei Lin and Yiwei Dai and Yuxin Wang and Wenhao Chai and Shang Zhou and Dariush Wahdany and Ziyu She and Jiaming Hu and Zhikang Dong and Yuxuan Zhu and Sasha Cui and Ahson Saiyed and Arinbjörn Kolbeinsson and Jesse Hu and Christopher Michael Rytting and Ryan Marten and Yixin Wang and Alex Dimakis and Andy Konwinski and Ludwig Schmidt},
      year={2026},
      eprint={2601.11868},
      archivePrefix={arXiv},
      primaryClass={cs.SE},
}

@misc{yang2025qwen3technicalreport,
      title={Qwen3 Technical Report}, 
      author={Team Qwen},
      year={2025},
      eprint={2505.09388},
      archivePrefix={arXiv},
      primaryClass={cs.CL},
}

@misc{kimiteam2026kimik25visualagentic,
      title={Kimi K2.5: Visual Agentic Intelligence}, 
      author={Team Kimi},
      year={2026},
      eprint={2602.02276},
      archivePrefix={arXiv},
      primaryClass={cs.CL},
}

@misc{li2026skillsbenchbenchmarkingagentskills,
      title={SkillsBench: Benchmarking How Well Agent Skills Work Across Diverse Tasks}, 
      author={Xiangyi Li and Wenbo Chen and Yimin Liu and Shenghan Zheng and Xiaokun Chen and Yifeng He and Yubo Li and Bingran You and Haotian Shen and Jiankai Sun and Shuyi Wang and Binxu Li and Qunhong Zeng and Di Wang and Xuandong Zhao and Yuanli Wang and Roey Ben Chaim and Zonglin Di and Yipeng Gao and Junwei He and Yizhuo He and Liqiang Jing and Luyang Kong and Xin Lan and Jiachen Li and Songlin Li and Yijiang Li and Yueqian Lin and Xinyi Liu and Xuanqing Liu and Haoran Lyu and Ze Ma and Bowei Wang and Runhui Wang and Tianyu Wang and Wengao Ye and Yue Zhang and Hanwen Xing and Yiqi Xue and Steven Dillmann and Han-chung Lee},
      year={2026},
      eprint={2602.12670},
      archivePrefix={arXiv},
      primaryClass={cs.AI},
}

@misc{gemini31pro_modelcard_2026,
  title        = {Gemini 3.1 Pro Model Card},
  author       = {{Google DeepMind}},
  year         = {2026},
  month        = {February},
  howpublished = {\url{https://storage.googleapis.com/deepmind-media/Model-Cards/Gemini-3-1-Pro-Model-Card.pdf}},
  note         = {Accessed: 2026-03-31}
}

@misc{openai_gpt54_thinking_2026,
  title        = {GPT-5.4 Thinking System Card},
  author       = {{OpenAI}},
  year         = {2026},
  howpublished = {\url{https://deploymentsafety.openai.com/gpt-5-4-thinking}},
  note         = {Accessed: 2026-03-31}
}

@article{anthropic2026opus46,
  title        = {Claude Opus 4.6 System Card},
  author       = {{Anthropic}},
  year         = {2026},
  url          = {https://www-cdn.anthropic.com/0dd865075ad3132672ee0ab40b05a53f14cf5288.pdf},
  note         = {System card describing model capabilities, evaluations, and safety assessments},
  institution  = {Anthropic}
}

@misc{openclaw2025github,
  author       = {Steinberger, Peter and OpenClaw Contributors},
  title        = {OpenClaw: Your own personal AI assistant},
  year         = {2025},
  howpublished = {\url{https://github.com/openclaw/openclaw}},
  note         = {GitHub repository},
}

@misc{openai_codex_2025,
  author       = {{OpenAI}},
  title        = {OpenAI Codex},
  year         = {2025},
  howpublished = {\url{https://openai.com/codex/}},
  note         = {Accessed: 2026-03-31}
}

@misc{agentskills2026,
  title        = {Agent Skills: A Simple, Open Format for Giving Agents New Capabilities},
  author       = {{Anthropic}},
  year         = {2026},
  howpublished = {\url{https://agentskills.io/home}},
  note         = {Accessed: 2026-03-31}
}

@misc{anthropic_claude_code_overview,
  author       = {{Anthropic}},
  title        = {Claude Code Documentation: Overview},
  year         = {2025},
  howpublished = {\url{https://code.claude.com/docs/en/overview}},
  note         = {Accessed: 2026-03-31}
}

@software{harbor2026,
  author = {{Harbor Framework Team}},
  month = jan,
  title = {{Harbor: A framework for evaluating and optimizing agents and models in container environments}},
  url = {https://github.com/harbor-framework/harbor},
  year = {2026}
}

@misc{wang2023voyageropenendedembodiedagent,
      title={Voyager: An Open-Ended Embodied Agent with Large Language Models}, 
      author={Guanzhi Wang and Yuqi Xie and Yunfan Jiang and Ajay Mandlekar and Chaowei Xiao and Yuke Zhu and Linxi Fan and Anima Anandkumar},
      year={2023},
      eprint={2305.16291},
      archivePrefix={arXiv},
      primaryClass={cs.AI},
}

@inproceedings{
nguyen2025dynasaur,
title={DynaSaur: Large Language Agents Beyond Predefined Actions},
author={Dang Nguyen and Viet Dac Lai and Seunghyun Yoon and Ryan A. Rossi and Handong Zhao and Ruiyi Zhang and Puneet Mathur and Nedim Lipka and Yu Wang and Trung Bui and Franck Dernoncourt and Tianyi Zhou},
booktitle={Second Conference on Language Modeling},
year={2025},
}

@inproceedings{
wang2025inducing,
title={Inducing Programmatic Skills for Agentic Tasks},
author={Zora Zhiruo Wang and Apurva Gandhi and Graham Neubig and Daniel Fried},
booktitle={Second Conference on Language Modeling},
year={2025},
}

@misc{shi2026evolvingprogrammaticskillnetworks,
      title={Evolving Programmatic Skill Networks}, 
      author={Haochen Shi and Xingdi Yuan and Bang Liu},
      year={2026},
      eprint={2601.03509},
      archivePrefix={arXiv},
      primaryClass={cs.AI},
}

@misc{wang2024agentworkflowmemory,
      title={Agent Workflow Memory}, 
      author={Zora Zhiruo Wang and Jiayuan Mao and Daniel Fried and Graham Neubig},
      year={2024},
      eprint={2409.07429},
      archivePrefix={arXiv},
      primaryClass={cs.CL},
}

@misc{mi2026procmemlearningreusableprocedural,
      title={ProcMEM: Learning Reusable Procedural Memory from Experience via Non-Parametric PPO for LLM Agents}, 
      author={Qirui Mi and Zhijian Ma and Mengyue Yang and Haoxuan Li and Yisen Wang and Haifeng Zhang and Jun Wang},
      year={2026},
      eprint={2602.01869},
      archivePrefix={arXiv},
      primaryClass={cs.AI},
}

@misc{zhao2024expelllmagentsexperiential,
      title={ExpeL: LLM Agents Are Experiential Learners}, 
      author={Andrew Zhao and Daniel Huang and Quentin Xu and Matthieu Lin and Yong-Jin Liu and Gao Huang},
      year={2024},
      eprint={2308.10144},
      archivePrefix={arXiv},
      primaryClass={cs.LG},
}

@misc{zheng2025skillweaverwebagentsselfimprove,
      title={SkillWeaver: Web Agents can Self-Improve by Discovering and Honing Skills}, 
      author={Boyuan Zheng and Michael Y. Fatemi and Xiaolong Jin and Zora Zhiruo Wang and Apurva Gandhi and Yueqi Song and Yu Gu and Jayanth Srinivasa and Gaowen Liu and Graham Neubig and Yu Su},
      year={2025},
      eprint={2504.07079},
      archivePrefix={arXiv},
      primaryClass={cs.AI},
}

@misc{yang2026autoskillexperiencedrivenlifelonglearning,
      title={AutoSkill: Experience-Driven Lifelong Learning via Skill Self-Evolution}, 
      author={Yutao Yang and Junsong Li and Qianjun Pan and Bihao Zhan and Yuxuan Cai and Lin Du and Jie Zhou and Kai Chen and Qin Chen and Xin Li and Bo Zhang and Liang He},
      year={2026},
      eprint={2603.01145},
      archivePrefix={arXiv},
      primaryClass={cs.AI},
}

@misc{alzubi2026evoskillautomatedskilldiscovery,
      title={EvoSkill: Automated Skill Discovery for Multi-Agent Systems}, 
      author={Salaheddin Alzubi and Noah Provenzano and Jaydon Bingham and Weiyuan Chen and Tu Vu},
      year={2026},
      eprint={2603.02766},
      archivePrefix={arXiv},
      primaryClass={cs.AI},
}

@misc{xia2026skillrlevolvingagentsrecursive,
      title={SkillRL: Evolving Agents via Recursive Skill-Augmented Reinforcement Learning}, 
      author={Peng Xia and Jianwen Chen and Hanyang Wang and Jiaqi Liu and Kaide Zeng and Yu Wang and Siwei Han and Yiyang Zhou and Xujiang Zhao and Haifeng Chen and Zeyu Zheng and Cihang Xie and Huaxiu Yao},
      year={2026},
      eprint={2602.08234},
      archivePrefix={arXiv},
      primaryClass={cs.LG},
}

@misc{wang2026reinforcementlearningselfimprovingagent,
      title={Reinforcement Learning for Self-Improving Agent with Skill Library}, 
      author={Jiongxiao Wang and Qiaojing Yan and Yawei Wang and Yijun Tian and Soumya Smruti Mishra and Zhichao Xu and Megha Gandhi and Panpan Xu and Lin Lee Cheong},
      year={2026},
      eprint={2512.17102},
      archivePrefix={arXiv},
      primaryClass={cs.AI},
}

@misc{jiang2026sokagenticskills,
      title={SoK: Agentic Skills -- Beyond Tool Use in LLM Agents}, 
      author={Yanna Jiang and Delong Li and Haiyu Deng and Baihe Ma and Xu Wang and Qin Wang and Guangsheng Yu},
      year={2026},
      eprint={2602.20867},
      archivePrefix={arXiv},
      primaryClass={cs.CR},
}

@misc{han2026sweskillsbenchagentskillsactually,
      title={SWE-Skills-Bench: Do Agent Skills Actually Help in Real-World Software Engineering?}, 
      author={Tingxu Han and Yi Zhang and Wei Song and Chunrong Fang and Zhenyu Chen and Youcheng Sun and Lijie Hu},
      year={2026},
      eprint={2603.15401},
      archivePrefix={arXiv},
      primaryClass={cs.SE},
}

@misc{zheng2026skillrouterskillroutingllm,
      title={SkillRouter: Skill Routing for LLM Agents at Scale}, 
      author={YanZhao Zheng and ZhenTao Zhang and Chao Ma and YuanQiang Yu and JiHuai Zhu and Yong Wu and Tianze Xu and Baohua Dong and Hangcheng Zhu and Ruohui Huang and Gang Yu},
      year={2026},
      eprint={2603.22455},
      archivePrefix={arXiv},
      primaryClass={cs.LG},
}

@misc{liang2026skillnetcreateevaluateconnect,
      title={SkillNet: Create, Evaluate, and Connect AI Skills}, 
      author={Yuan Liang and Ruobin Zhong and Haoming Xu and Chen Jiang and Yi Zhong and Runnan Fang and Jia-Chen Gu and Shumin Deng and Yunzhi Yao and Mengru Wang and Shuofei Qiao and Xin Xu and Tongtong Wu and Kun Wang and Yang Liu and Zhen Bi and Jungang Lou and Yuchen Eleanor Jiang and Hangcheng Zhu and Gang Yu and Haiwen Hong and Longtao Huang and Hui Xue and Chenxi Wang and Yijun Wang and Zifei Shan and Xi Chen and Zhaopeng Tu and Feiyu Xiong and Xin Xie and Peng Zhang and Zhengke Gui and Lei Liang and Jun Zhou and Chiyu Wu and Jin Shang and Yu Gong and Junyu Lin and Changliang Xu and Hongjie Deng and Wen Zhang and Keyan Ding and Qiang Zhang and Fei Huang and Ningyu Zhang and Jeff Z. Pan and Guilin Qi and Haofen Wang and Huajun Chen},
      year={2026},
      eprint={2603.04448},
      archivePrefix={arXiv},
      primaryClass={cs.AI},
}

@misc{schmotz2026skillinjectmeasuringagentvulnerability,
      title={Skill-Inject: Measuring Agent Vulnerability to Skill File Attacks}, 
      author={David Schmotz and Luca Beurer-Kellner and Sahar Abdelnabi and Maksym Andriushchenko},
      year={2026},
      eprint={2602.20156},
      archivePrefix={arXiv},
      primaryClass={cs.CR},
}

@misc{zhou2026mementoskillsletagentsdesign,
      title={Memento-Skills: Let Agents Design Agents}, 
      author={Huichi Zhou and Siyuan Guo and Anjie Liu and Zhongwei Yu and Ziqin Gong and Bowen Zhao and Zhixun Chen and Menglong Zhang and Yihang Chen and Jinsong Li and Runyu Yang and Qiangbin Liu and Xinlei Yu and Jianmin Zhou and Na Wang and Chunyang Sun and Jun Wang},
      year={2026},
      eprint={2603.18743},
      archivePrefix={arXiv},
      primaryClass={cs.AI},
}

@misc{shinn2023reflexionlanguageagentsverbal,
      title={Reflexion: Language Agents with Verbal Reinforcement Learning}, 
      author={Noah Shinn and Federico Cassano and Edward Berman and Ashwin Gopinath and Karthik Narasimhan and Shunyu Yao},
      year={2023},
      eprint={2303.11366},
      archivePrefix={arXiv},
      primaryClass={cs.AI},
}

@misc{yao2024retroformerretrospectivelargelanguage,
      title={Retroformer: Retrospective Large Language Agents with Policy Gradient Optimization}, 
      author={Weiran Yao and Shelby Heinecke and Juan Carlos Niebles and Zhiwei Liu and Yihao Feng and Le Xue and Rithesh Murthy and Zeyuan Chen and Jianguo Zhang and Devansh Arpit and Ran Xu and Phil Mui and Huan Wang and Caiming Xiong and Silvio Savarese},
      year={2024},
      eprint={2308.02151},
      archivePrefix={arXiv},
      primaryClass={cs.CL},
}

@misc{suzgun2025dynamiccheatsheettesttimelearning,
      title={Dynamic Cheatsheet: Test-Time Learning with Adaptive Memory}, 
      author={Mirac Suzgun and Mert Yuksekgonul and Federico Bianchi and Dan Jurafsky and James Zou},
      year={2025},
      eprint={2504.07952},
      archivePrefix={arXiv},
      primaryClass={cs.LG},
}

@misc{ouyang2026reasoningbankscalingagentselfevolving,
      title={ReasoningBank: Scaling Agent Self-Evolving with Reasoning Memory}, 
      author={Siru Ouyang and Jun Yan and I-Hung Hsu and Yanfei Chen and Ke Jiang and Zifeng Wang and Rujun Han and Long T. Le and Samira Daruki and Xiangru Tang and Vishy Tirumalashetty and George Lee and Mahsan Rofouei and Hangfei Lin and Jiawei Han and Chen-Yu Lee and Tomas Pfister},
      year={2026},
      eprint={2509.25140},
      archivePrefix={arXiv},
      primaryClass={cs.AI},
}

@misc{zhang2026liveevoonlineevolutionagentic,
      title={Live-Evo: Online Evolution of Agentic Memory from Continuous Feedback}, 
      author={Yaolun Zhang and Yiran Wu and Yijiong Yu and Qingyun Wu and Huazheng Wang},
      year={2026},
      eprint={2602.02369},
      archivePrefix={arXiv},
      primaryClass={cs.AI},
}

@misc{zhou2025mementofinetuningllmagents,
      title={Memento: Fine-tuning LLM Agents without Fine-tuning LLMs}, 
      author={Huichi Zhou and Yihang Chen and Siyuan Guo and Xue Yan and Kin Hei Lee and Zihan Wang and Ka Yiu Lee and Guchun Zhang and Kun Shao and Linyi Yang and Jun Wang},
      year={2025},
      eprint={2508.16153},
      archivePrefix={arXiv},
      primaryClass={cs.LG},
}

@misc{zhang2026memskilllearningevolvingmemory,
      title={MemSkill: Learning and Evolving Memory Skills for Self-Evolving Agents}, 
      author={Haozhen Zhang and Quanyu Long and Jianzhu Bao and Tao Feng and Weizhi Zhang and Haodong Yue and Wenya Wang},
      year={2026},
      eprint={2602.02474},
      archivePrefix={arXiv},
      primaryClass={cs.CL},
}

@misc{yan2026tidetrajectorybaseddiagnosticevaluation,
      title={TIDE: Trajectory-based Diagnostic Evaluation of Test-Time Improvement in LLM Agents}, 
      author={Hang Yan and Xinyu Che and Fangzhi Xu and Qiushi Sun and Zichen Ding and Kanzhi Cheng and Jian Zhang and Tao Qin and Jun Liu and Qika Lin},
      year={2026},
      eprint={2602.02196},
      archivePrefix={arXiv},
      primaryClass={cs.AI},
}

@misc{fang2025comprehensivesurveyselfevolvingai,
      title={A Comprehensive Survey of Self-Evolving AI Agents: A New Paradigm Bridging Foundation Models and Lifelong Agentic Systems}, 
      author={Jinyuan Fang and Yanwen Peng and Xi Zhang and Yingxu Wang and Xinhao Yi and Guibin Zhang and Yi Xu and Bin Wu and Siwei Liu and Zihao Li and Zhaochun Ren and Nikos Aletras and Xi Wang and Han Zhou and Zaiqiao Meng},
      year={2025},
      eprint={2508.07407},
      archivePrefix={arXiv},
      primaryClass={cs.AI},
}
\bibliographystyle{colm2026_conference}

\appendix
\section{Skill Search Engine Details}
\label{sec:retrieval_details}

\paragraph{Skill index construction.}
We index the full collection of 34,198 skills with two complementary representations.
For each skill, we extract: \ding{172} \emph{metadata}, formed by concatenating the skill's name and description, and \ding{173} \emph{full content}, the body of the \texttt{SKILL.md} file.
We filter the collection to skills with permissive licenses (MIT and Apache-2.0).

For sparse retrieval, we build an SQLite FTS5 full-text search index over the metadata fields.
BM25 ranking uses field weights of 10 for name, 5 for description, and 5 for full content (when the content field is included in the index).
The FTS5 index supports standard query syntax including prefix matching, phrase queries, and boolean operators.

For dense retrieval, we compute embeddings using Qwen3-Embedding-4B.
At query time, we prepend the query instruction ``\texttt{Find skills matching this query:}'' before encoding.

\paragraph{Search tools.}
We implement three search endpoints exposed to the agent via an HTTP server:
\begin{itemize}[leftmargin=1.5em]
    \item \textbf{Keyword search} (\texttt{/keyword}): BM25-based retrieval over the FTS5 index.
    \item \textbf{Semantic search} (\texttt{/semantic}): Dense embedding cosine similarity.
    \item \textbf{Hybrid search} (\texttt{/hybrid}): Combines keyword and semantic results using Reciprocal Rank Fusion (RRF). Specifically, the RRF score for a skill is $\sum_{s} w_s / (k + r_s)$, where $r_s$ is the rank in search method $s$, $w_s$ is the method weight (default 0.5 for both keyword and semantic), and $k=60$ is the fusion constant. The keyword and semantic weights are configurable per query by the agent.
\end{itemize}
A separate \texttt{/detail} endpoint retrieves the full \texttt{SKILL.md} content for any skill given its identifier.

For agentic search variants that include the full content index (\emph{hybrid w/ content} in Table~\ref{tab:retrieval}), the semantic similarity score is computed as a weighted average of metadata and content embedding similarities: $(1 - w) \cdot \text{sim}_{\text{meta}} + w \cdot \text{sim}_{\text{content}}$.

To select the content weight $w$ and BM25 content field weight, we sweep over candidate values using synthetic queries: we prompt a model to generate 1-3 short search queries per task from the task instruction, then use these queries for direct (non-agentic) search and measure Recall@5 against the curated skills.
The best-performing configuration uses a BM25 content field weight of 5 and a semantic content weight of $w = 0.05$.

\paragraph{Agentic search protocol.}
In the agentic search setting, the agent is provided with a \emph{finding-skills} skill that describes the search API and a structured workflow for discovering relevant skills.
The full content of this skill is shown below.

\begin{tcolorbox}[breakable, enhanced, title={\small\bfseries Finding-Skills Skill (provided to the agent for retrieval)}, colback=gray!5, colframe=gray!50, fonttitle=\bfseries\small, left=2pt, right=2pt, top=2pt, bottom=2pt]
\begin{lstlisting}[style=promptStyle]
# Finding Skills

Searches a local index of agent skills to find the most relevant ones for a given task. Skills are pre-downloaded - no installation needed.

## When to Use

- Starting a new task that may benefit from specialized skills
- Looking for best practices, patterns, or workflows for a specific domain
- Wanting to find tools or templates for a task (testing, deployment, design, etc.)

## Search API

Three search endpoints are available. Use `curl -s` to query them.

### Keyword search

Best for exact term matching when you know specific skill names or technologies.

```bash
curl -s "http://localhost:8742/keyword?q=QUERY&top_k=10"
```

Supports FTS5 syntax:
- Prefix: `react*`
- Phrase: `"code review"`
- Boolean: `react OR vue`

### Semantic search

Best for conceptual queries where you describe what you need in natural language.

```bash
curl -s "http://localhost:8742/semantic?q=QUERY&top_k=10"
```

Example: `q=help me build and deploy containerized applications`

### Hybrid search

Combines keyword and semantic search with reciprocal rank fusion. Best general-purpose option.

```bash
curl -s "http://localhost:8742/hybrid?q=QUERY&top_k=10&keyword_weight=0.5&semantic_weight=0.5"
```

- `keyword_weight`: How much BM25 keyword matches contribute to the final ranking (default: 0.5)
- `semantic_weight`: How much semantic similarity contributes to the final ranking (default: 0.5)

### Response format

All search endpoints return a JSON array. Each result contains:

- `name`: skill name (may be duplicated across authors)
- `description`: what the skill does
- `skill_md_snippet`: first 100 words of the skill's documentation
- `skill_id`: unique identifier (author--name format), use with the detail endpoint
- `github_stars`: popularity of the source repository
- `score`: relevance score. For keyword search, more negative = better match. For semantic search, 0-1 where higher = better match. Hybrid search returns `rrf_score` instead (higher = better match).

### Skill detail

Fetches full metadata and complete SKILL.md content for a skill. Pass the `skill_id` from search results.

```bash
curl -s "http://localhost:8742/detail/SKILL_ID"
```

## Workflow

### Step 1: Analyze the task

Break the task into concrete sub-tasks. For example, "build a REST API with auth and tests" becomes:
- Design API endpoints and routing
- Implement authentication
- Write tests

### Step 2: Search for each sub-task

For each sub-task, run search queries to find skills.

```bash
# Sub-task: implement authentication
curl -s "http://localhost:8742/hybrid?q=implement+authentication+JWT&top_k=10"
# Keyword search
curl -s "http://localhost:8742/keyword?q=JWT&top_k=10"

# Sub-task: write tests
curl -s "http://localhost:8742/hybrid?q=writing+unit+tests&top_k=10"
```

Refine queries if initial results are too broad or miss the mark. Try different phrasing or switch between keyword and semantic search.

### Step 3: Review and select

From the search results, select 10 skills total across all sub-tasks. Prioritize:
- High relevance to the target task
- Higher `github_stars` when multiple skills cover the same topic
- Skills with informative `skill_md_snippet` content

If needed, fetch full details of a skill to confirm relevance:

```bash
curl -s "http://localhost:8742/detail/SKILL_ID"
```

### Step 4: Record results

Record the selected skills as a structured list:

```
## Found Skills

- **[skill-name]** (skill_id: [skill_id]) - [one-line summary from description]
- **[skill-name]** (skill_id: [skill_id]) - [one-line summary from description]
- ...
```
\end{lstlisting}
\end{tcolorbox}

\section{Experiment Details}
\label{sec:experiment_details}

\paragraph{Benchmarks.}
We evaluate on two benchmarks:
\begin{itemize}[leftmargin=1.5em]
    \item \textsc{SkillsBench}~\citep{li2026skillsbenchbenchmarkingagentskills}: We use 84 tasks, excluding 3 tasks with known environment or verifier issues: \texttt{mhc-layer-impl}, \texttt{scheduling-email-assistant}, and \texttt{fix-visual-stability}.
    \item \textsc{Terminal-Bench 2.0}~\citep{merrill2026terminalbenchbenchmarkingagentshard}: We use all 89 tasks.
\end{itemize}

\paragraph{Models and agent harnesses.}
We evaluate three models, each paired with its native agent harness:
\begin{itemize}[leftmargin=1.5em]
    \item \texttt{Claude Opus 4.6}~\citep{anthropic2026opus46} with Claude Code v2.1.19.
    \item \texttt{Kimi K2.5}~\citep{kimiteam2026kimik25visualagentic} with Terminus-2~\citep{harbor2026} (max input tokens: 253,952), served locally via SGLang~\citep{zheng2024sglangefficientexecutionstructured}.
    \item \texttt{Qwen/Qwen3.5-397B-A17B-FP8}~\citep{yang2025qwen3technicalreport} with Qwen-Code v0.12.3, served locally via SGLang.
\end{itemize}

\paragraph{Evaluation protocol.}
All experiments are run in isolated Docker containers provided by each task using the Harbor framework~\citep{harbor2026}.
Each task is run 3 times, and results are evaluated using the benchmark's automated verifiers.
On \textsc{SkillsBench}, we use a timeout multiplier of 1.5$\times$ the default task timeout for all three models.
On \textsc{Terminal-Bench 2.0}, we use a 2$\times$ timeout multiplier for \texttt{Kimi K2.5} and \texttt{Qwen3.5} to account for the lower inference speed of local serving, while keeping the original timeout (1$\times$) for \texttt{Claude Opus 4.6}.

\section{Skill Refinement Details}
\label{sec:refinement_details}

\subsection{Query-Specific Refinement}
\label{sec:query_specific_details}

Query-specific refinement runs inside the task's own Docker environment, giving the agent access to the task's data, libraries, and tools. However, the agent does not have access to the ground-truth verifier and needs to self-evaluate the correctness of a trajectory. We limit the refinement to a single iteration.

The full instruction prompt given to the refinement agent is shown below:

\begin{tcolorbox}[breakable, enhanced, title={\small\bfseries Query-Specific Refinement Prompt}, colback=gray!5, colframe=gray!50, fonttitle=\bfseries\small, left=2pt, right=2pt, top=2pt, bottom=2pt]
\begin{lstlisting}[style=promptStyle]
You are a skill refinement agent. Your goal is to attempt the target task using the retrieved skills, observe which parts of the skills help and which don't, and then create improved refined skills based on that experience.

## Your Task

### Phase 1: Understand the task and skills

1. Read the task description in /root/task_instruction.md to understand what the task requires.
2. Read ALL the retrieved skills in /root/retrieved_skills/. Each subdirectory contains a skill with a SKILL.md and possibly supporting files (scripts, references, etc.).

### Phase 2: Attempt the task using the retrieved skills

3. Try to solve the task while actively consulting the retrieved skills. This is the most important step. As you work through the task:
   - Refer to the retrieved skills for guidance, code snippets, API patterns, and domain knowledge.
   - When a skill suggests an approach, try it. Note whether it works, partially works, or is wrong.
   - When you get stuck, check if any skill covers the issue. Note gaps where no skill helps.
   - Keep track of which specific parts of which skills were useful, misleading, or irrelevant.

   IMPORTANT: If you delegate any part of the exploration to a subagent, you MUST give that subagent access to the retrieved skills at /root/retrieved_skills/ and instruct it to consult them during its work. The goal is to test the skills in practice, not to solve the task from scratch independently.

### Phase 3: Reflect and create refined skills

4. Based on your experience attempting the task with the retrieved skills, reflect on:
   - Which skills or parts of skills were directly useful?
   - Which skills had errors, outdated information, or misleading guidance?
   - What knowledge was missing that you had to figure out on your own?
   - What would have made the task easier if you had known it upfront?

5. Use the skill-creator skill at {agent_skills_path}/skill-creator/ as guidance for creating and writing skills.

6. Create refined skills that incorporate what you learned. The refined skills should:
   - Keep the parts that actually worked when you tried them.
   - Fix or remove parts that were wrong or misleading.
   - Add knowledge you discovered during exploration that was missing from the original skills.
   - Combine related information from multiple skills into coherent, task-appropriate guides.

## Important Guidelines

- **Focus on this task only.** You are preparing skills specifically for this given task. There is no need to create additional test queries - just test and evaluate against the task in /root/task_instruction.md. You do not have access to the ground-truth verifier; judge quality based on your own knowledge and exploration of the task.
- **Single iteration of improvement.** Do one round of exploration and refinement - do not iterate multiple times.
- **No user interaction.** Do not ask any questions. Self-explore the target task and create improved skills based on your exploration trajectory.
- **Compose, don't copy.** The refined skills do not need to cover all information in the retrieved skills. Instead, extract and compose the useful, relevant parts and combine them into coherent skills. There is no limit on the number of skills you create - you can create more or fewer than the number of retrieved skills. Focus on quality and relevance.

## Output

Save your refined skills to {refined_skills_path}/. Each skill should be in its own subdirectory with a SKILL.md file (and optional supporting files like scripts or references):

```
{refined_skills_path}/
+-- skill-name-1/
|   +-- SKILL.md
|   +-- (optional supporting files)
+-- skill-name-2/
|   +-- SKILL.md
|   +-- (optional supporting files)
+-- ...
```
\end{lstlisting}
\end{tcolorbox}

\subsection{Query-Agnostic Refinement}
\label{sec:query_agnostic_details}

Query-agnostic refinement improves each skill independently without knowledge of any target task. Each skill is refined in a minimal Docker container (Ubuntu 24.04 with Python).

The instruction prompt given to the refinement agent is:

\begin{tcolorbox}[breakable, enhanced, title={\small\bfseries Query-Agnostic Refinement Prompt}, colback=gray!5, colframe=gray!50, fonttitle=\bfseries\small, left=2pt, right=2pt, top=2pt, bottom=2pt]
\begin{lstlisting}[style=promptStyle]
You are a skill improvement agent. Your job is to improve a single skill.

The skill to improve is at /root/skill_to_improve/ (contains SKILL.md and possibly supporting files). A guide on how to create and improve skills (skill-creator) is available in your skills.

Read the skill to improve and the skill-creator guide. Follow the skill-creator methodology to generate sample test queries, then evaluate the skill using A/B testing as described in the guide. Finally, improve the skill based on what you find.

## Important Guidelines

- **No user interaction.** Work autonomously.
- **Single iteration.** One round of evaluation and improvement - do not loop multiple times.

## Output

Save the improved skill to {refined_skill_path}/. The skill should have a SKILL.md file and optional supporting files:

```
{refined_skill_path}/
+-- SKILL.md
+-- (optional supporting files like scripts/, references/, assets/)
```
\end{lstlisting}
\end{tcolorbox}

\section{LLM-as-Judge for Skill Coverage}
\label{sec:coverage_details}

To assess the relevance and coverage of retrieved skill sets (Table~\ref{tab:coverage}), we use \texttt{GPT-5.4} as an LLM judge.
For each task, the judge receives the task instruction and the full content of all retrieved skills (including \texttt{SKILL.md} and helper files, truncated to 400K characters per skill and 2M characters total), and is asked to rate overall coverage.
The system prompt is:

\begin{tcolorbox}[breakable, enhanced, title={\small\bfseries LLM Judge System Prompt}, colback=gray!5, colframe=gray!50, fonttitle=\bfseries\small, left=2pt, right=2pt, top=2pt, bottom=2pt]
\begin{lstlisting}[style=promptStyle]
You are an expert evaluator assessing how well a set of skill documents collectively covers a specific task. A "skill" is a reusable knowledge document (with optional helper scripts/references) that an AI agent can consult while working on a task.

You will be given a task instruction and a set of skills. Evaluate how well the skills TOGETHER cover what is needed to complete the task.

Rate overall coverage on this scale:
  5 = Complete coverage - the skills together cover all steps and aspects needed to solve the task. An agent with these skills has everything it needs.
  4 = High coverage - the skills cover most aspects of the task, but minor gaps remain that the agent would need to figure out on its own.
  3 = Moderate coverage - the skills cover some key aspects but miss significant parts of the task. The agent would need substantial independent work.
  2 = Low coverage - the skills touch on the topic but miss most of the task's specific needs. Only marginally helpful.
  1 = No coverage - nothing in the skills is relevant to the task.

Respond with ONLY a JSON object, no explanation:
{"score": <1|2|3|4|5>, "covered": "<what the skills cover>", "gaps": "<what is missing>"}
\end{lstlisting}
\end{tcolorbox}

\end{document}